\documentclass[9pt]{article}
\usepackage[utf8]{inputenc}
\usepackage[T1]{fontenc}
\usepackage{microtype}
\usepackage[margin=1in]{geometry}
\usepackage{times}
\usepackage{graphicx} 
\usepackage{caption}
\usepackage[hidelinks]{hyperref}
\usepackage{placeins}
\usepackage{amsmath,amssymb}
\usepackage{authblk}
\usepackage{listings}
\title{Fostering Robots: A Governance-First Conceptual Framework for Domestic, Curriculum-Based Trajectory Collection}
\author[1]{Federico Pablo-Mart'{i}}
\author[1]{Carlos Mir Fernández}

\affil[1]{Complex Systems in Social Sciences Group (SCCS), University of Alcal'{a}}

\date{} 
\begin{document}
\maketitle
\begin{abstract}
We propose a conceptual, empirically testable framework for 'Robot Fostering'—a curriculum-driven, governance-first approach to domestic robot deployments, emphasizing long-term, curated interaction trajectories. We formalize trajectory quality with quantifiable metrics and evaluation protocols aligned with EU-grade governance standards, delineating a low-resource empirical roadmap to enable rigorous validation through future pilot studies. \end{abstract} \section*{Main Text} \section{From guide-dog puppy raisers'' to robot fostering}
A core lesson from recent work on agentic AI is that \emph{agency scales with the quality and structure of experience}, not with indiscriminate volume. LIMI (Less Is More for Agency'') fine-tunes a capable model with just 78 carefully curated, long trajectories---coding and research workflows---achieving 73.5\% on AgencyBench and outperforming models trained on orders-of-magnitude more data \cite{Xiao2025LIMI}. We argue that robotics can---and should---exploit the same principle. Our intent is expressly conceptual: a blueprint that specifies methods, governance, and testable predictions without claiming empirical superiority today. We propose \emph{Robot Fostering}: a governance-first model that places domestic robots for time-limited, structured exposure curricula'' in ordinary homes, modelled on the guide-dog puppy-raising paradigm. Families host robots under clear protocols; exposure follows staged goals (play $\rightarrow$ family $\rightarrow$ school analogues); data capture is privacy-preserving; evaluation emphasizes long, ecologically valid trajectories rather than isolated snippets.
Distribution feasibility has precedents in Willow Garage's PR2 Beta Program, which successfully placed 11 PR2s across international labs to accelerate capability development \cite{IEEE2010PR2,ROS2010PR2,UniBremenPR2}. Governance is catalyzed by the EU AI Act---with obligations for General-Purpose AI effective 2 Aug 2025---and by emerging AI management and transparency standards (ISO/IEC 42001; ISO/IEC 23894; IEEE 7001) \cite{EC_AIAct,ISO42001,ISO23894,IEEE7001}. The iRobot test-unit leak is a cautionary case underscoring why a governance-first design with strict data minimization, on-device redaction, and contractor controls is non-negotiable \cite{BI_RoombaLeak}.
\paragraph{Scope and nature of contribution.}
This is a conceptual Perspective. We do not claim empirical superiority of curricula over unstructured exposure in embodied robotics. Instead, we (i) formalize trajectory value, (ii) derive testable predictions, and (iii) specify governance and evaluation protocols enabling future low-resource pilots. All results are therefore conditional and falsifiable.
\section{The agency-efficiency principle for embodied robots}
LIMI demonstrates that few, long, richly annotated trajectories can train agentic competence more efficiently than massive but shallow corpora \cite{Xiao2025LIMI}. In robotics, long-horizon, multi-step domestic trajectories encode precisely what generalization needs: task decomposition, tool use, error recovery, and social navigation. A fostering program operationalizes these signals by design: ecology (real homes), narrative structure (intent$\rightarrow$plan$\rightarrow$act$\rightarrow$feedback$\rightarrow$re-plan$\rightarrow$closure), metacognitive summaries, and tool-enabled action. This mirrors how participatory/citizen sensing achieves value with ecologically valid data from real environments \cite{Coulson2021,MS_Toolkit}.
\paragraph{Formalizing trajectory value.}
Let $\tau={(s_t,a_t,o_t,\alpha_t)}_{t=1}^T$ be an embodied trajectory with annotations $\alpha_t$ (plan steps, errors, replans, reflections). We define its informational richness as
$$\mathcal{R}(\tau)= I(A;O\,|\,S) + H(\text{rare events}) + \lambda\,\mathrm{Delib}(\alpha),$$
where $I(A;O|S)$ captures controllability, $H(\cdot)$ the coverage of low-frequency phenomena (e.g., glare, occlusions, pet interference), and $\mathrm{Delib}(\alpha)$ the density of metacognitive markers. Practical proxies include MDL/compressibility, kNN novelty over feature embeddings, and the rate of error--recovery segments.
\paragraph{Decision-theoretic framing.}
Tasks can be cast as (Partially Observable) Markov Decision Processes \cite{Puterman1994,SuttonBarto2018}, where trajectory design is an \emph{information-gathering policy} optimizing downstream generalization under token budgets. Annotations can reference a lightweight ontology (e.g., OWL 2 classes for rooms, affordances, hazards) to standardize $\alpha_t$ and enable cross-home reasoning \cite{W3C_OWL2}.
\textbf{Agency-efficiency conjecture (embodied).}
\emph{For a fixed token budget, curricula that maximize $\sum \mathcal{R}(\tau)$ over long trajectories yield higher generalization on hold-out homes than i.i.d.\ short snippets.}
\section{A curricular exposure model (Edu-LIMI)}
We cast the fostering curriculum in three staged educational'' phases: \textbf{Play} (single-room navigation, soft obstacles), \textbf{Family} (multi-room patrols, portals and elevators, variable illumination), and \textbf{School} (small household projects, tool-enabled actions). Each phase prescribes checklists, environmental variability, and error-inducing events, with safety gates before promotion. An annotation schema aligned to a home-robot ontology (rooms, fixtures, objects, social zones) improves comparability and reduces label drift. The goal is transfer: improved success@K and reduced interventions in unseen homes. This curriculum is compatible with low-cost platforms and with data-mule'' patterns validated in field robotics \cite{Bhadauria2011, Tekdas2008, BhadauriaWiley}.
\begin{figure}[htbp]
\centering
\includegraphics[width=\linewidth]{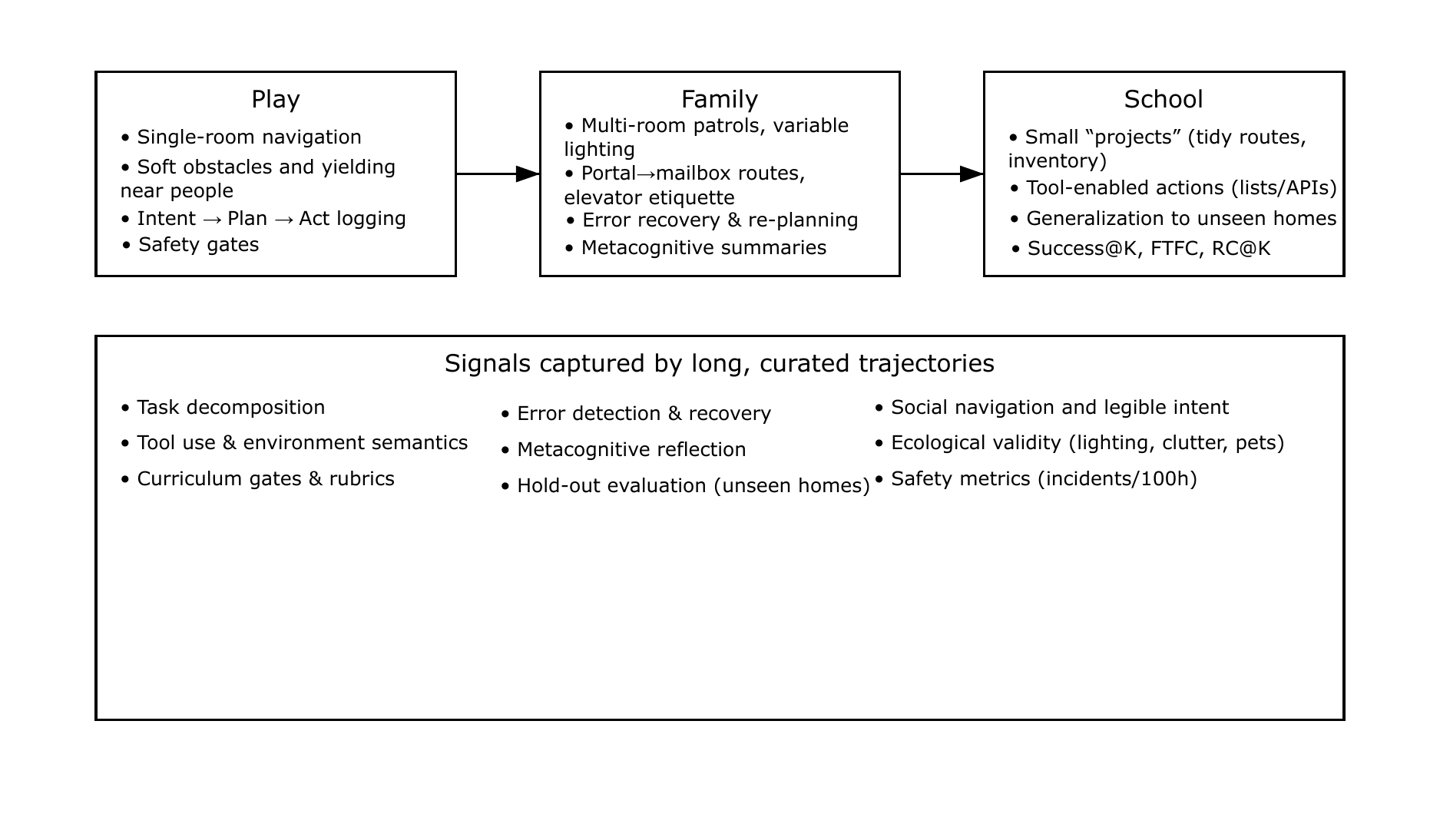}
\caption{Edu-LIMI exposure curriculum: three phases (Play, Family, School) mapped to agentic sub-skills; arrows indicate progression gates. Long, curated trajectories capture task decomposition, error recovery, social navigation, tool use, and metacognitive reflection.}
\end{figure}
\FloatBarrier
\section{Governance first: aligning with EU law and international standards}
A fostering program must be privacy-by-design and compliance-ready from day zero. The EU AI Act entered into force on 1 Aug 2024; prohibitions and AI literacy obligations apply from 2 Feb 2025; GPAI obligations apply from 2 Aug 2025; high-risk product rules have extended transitions to 2027 \cite{EC_AIAct,AIActTimeline,ReutersAIAct}. Establish an AI Management System (ISO/IEC 42001) and AI risk management (ISO/IEC 23894) integrated with DPIAs and vendor oversight \cite{ISO42001,ISO23894}; adopt IEEE 7001 levels for stakeholder-appropriate transparency \cite{IEEE7001}; map use to ISO 13482 for personal care robot safety and borrow AGV safety practices for motion in shared spaces \cite{ISO13482}.
\paragraph{Multi-framework alignment.}
Beyond EU law, we map roles and risks to \emph{NIST AI RMF 1.0}\cite{NIST_RMF} to harmonize vocabulary across jurisdictions, while \emph{GDPR Art.\ 35}\cite{GDPR35} (DPIA) triggers remain the binding basis in the EU. We propose redaction audits using synthetic fixtures (faces/plates) to estimate recall/precision and leakage rates.
\section{Evidence from adjacent deployments: distribution and acceptance}
Distribution feasibility is not hypothetical: PR2 Beta demonstrated that placing advanced robots across sites accelerates capability and software ecosystem growth \cite{IEEE2010PR2,ROS2010PR2,UniBremenPR2}. Meanwhile, autonomous delivery robots (ADRs) have accumulated empirical evidence on acceptance and externalities; systematic reviews and discrete-choice studies identify context, appearance, operations, and safety as key determinants of public attitudes \cite{Alverhed2024,Pani2020,Starship2025}.
\paragraph{Domain mismatch: ADRs vs homes.}
Evidence from sidewalk delivery robots informs legibility and speed control, but not household privacy or family dynamics. We therefore treat ADR findings as partial priors, and introduce home-specific safeguards (visitor notices, opt-out capture, transparent complaint logs).
\begin{figure}[htbp]
\centering
\includegraphics[width=\linewidth]{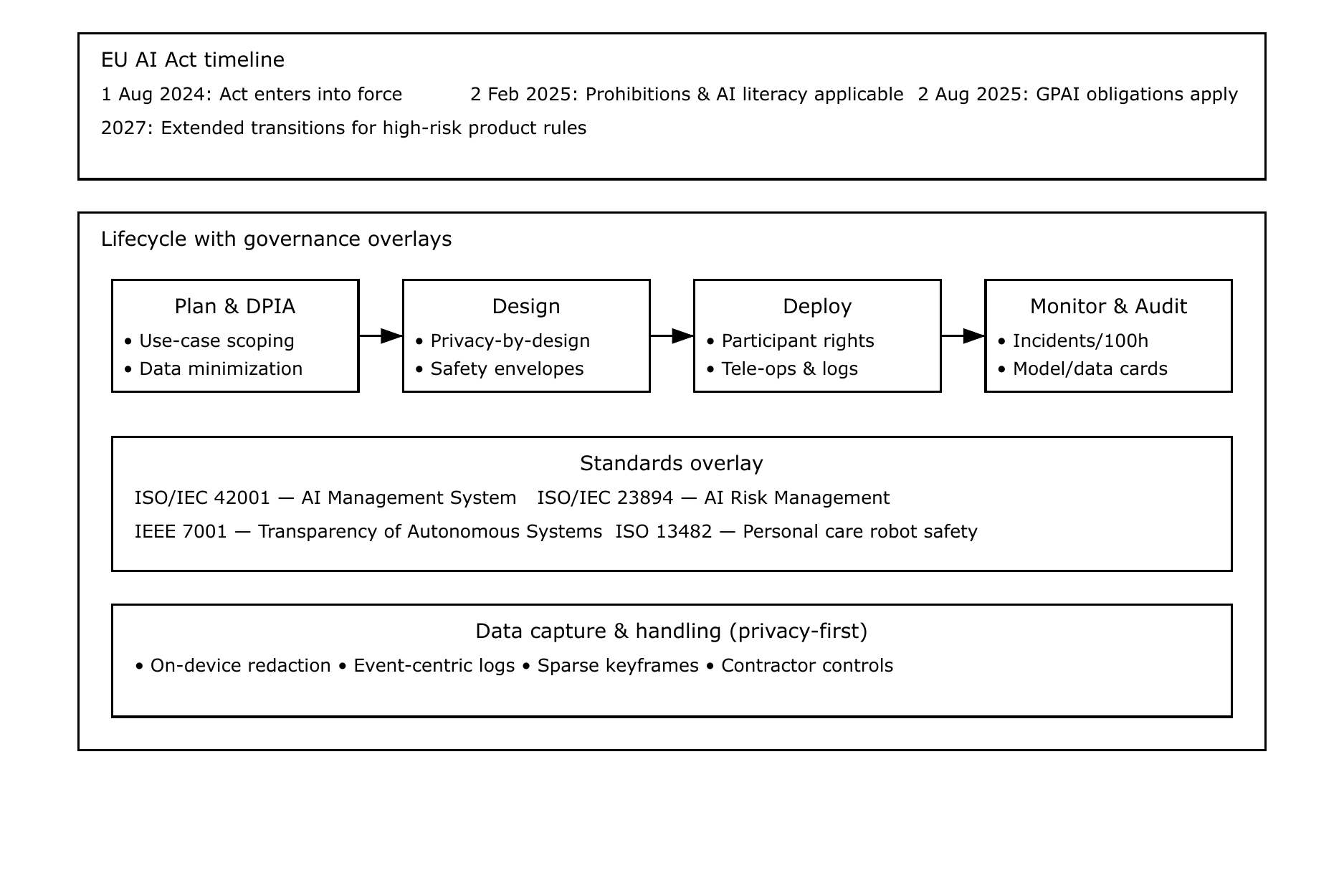}
\caption{Governance stack aligned to EU AI Act timeline and international standards: lifecycle phases (Plan \& DPIA, Design, Deploy, Monitor \& Audit) overlaid with ISO/IEC 42001 (AIMS), ISO/IEC 23894 (risk), IEEE 7001 (transparency), and ISO 13482 (personal-care robot safety).}
\end{figure}
\FloatBarrier
\section{A minimal research program (12--24 months; low-resource)}
We outline three complementary experiments: (i) Curriculum vs.\ control (random unstructured exposure vs.\ information-gain active placement vs.\ curricular exposure), (ii) Tool-free vs.\ tool-enabled evaluation (mirroring LIMI's with/without CLI), and (iii) Hold-out homes and shift tests. Metrics include success@3, FTFC@R, replan-cost@3, incidents/100h, post-encroachment time at doorways, and standardized bystander/user ratings adapted from ADR studies. For task taxonomies, results can be aligned post hoc to household benchmarks such as BEHAVIOR-1K (without running them), to connect constructs across communities \cite{Behavior1K}.
\paragraph{Illustrative simulation (toy).}
To demonstrate testability without CAPEX, we replicated a simple graph-based simulation (three phases with safety gates; edge error probabilities set synthetically). With per-phase success rates $(p_{\text{Play}},p_{\text{Family}},p_{\text{School}})\approx(0.85,0.60,0.70)$, the chained success for a full curriculum episode is $p\approx 0.85\times 0.60\times 0.70\approx 0.36$. This \emph{illustrative} result highlights that marginal improvements at any gate compound multiplicatively, justifying progressive curricula. The toy model is deliberately minimal and serves only to show how hypotheses (e.g., FTFC gains) can be quantified prior to physical pilots.
\paragraph{Power (correct and transparent).}
We compute sample sizes using standard frequentist power analyses. For continuous outcomes (e.g., FTFC, mean RC@K), a two-arm independent-samples t-test with $\alpha=0.05$, $1-\beta=0.80$, and a medium effect size (Cohen's $d=0.5$) requires \textbf{$n=64$ homes per arm}. For a balanced three-arm ANOVA with medium effect (Cohen's $f=0.25$), the total sample is \textbf{$n=158$} ($\approx 53$ per arm). For proportion outcomes (success@3), detecting an increase from $0.50$ to $0.65$ at $\alpha=0.05$ and $1-\beta=0.80$ requires \textbf{$n=167$ per arm} (two-proportion test). Where resources constrain recruitment, we pre-register (i) paired/repeated-measures designs to leverage within-home correlation, and (ii) group-sequential monitoring (e.g., Pocock boundaries) to preserve Type I error while allowing early stopping. Code snippets are provided in Appendix~A.
\section{Ethics, scope, and civil-only commitments}
\paragraph{Data ethics and curation biases.}
Beyond compliance, critical data-governance work warns about structural biases and documentation gaps in dataset curation \cite{GebruDatasheets,BenderDataStatements,BenderParrots}. We therefore pair GDPR/NIST-style controls with dataset \emph{datasheets} and \emph{data statements} to surface sampling, consent, and annotation decisions.
The fostering program is restricted to peaceful, civilian uses; dual-use risks are mitigated by narrow task definitions, transparency, and open oversight. To avoid socio-economic skew, use stratified sampling and non-coercive compensation; provide participant dashboards to inspect, audit, and delete records.
\section*{Limitations and falsifiability}
Our claims are conceptual. The embodied transfer of the agency-efficiency principle is a conjecture: if, under a fixed token budget, curricular exposure fails to improve FTFC/RC@3 on hold-out homes, H1 is falsified. Social viability is also uncertain; high opt-out or complaint rates would invalidate our feasibility assumptions. Privacy-by-design requires auditable on-device redaction and sealed annotation supply-chains; absent these, leakage risks remain.
\section*{Why this matters now}
As agentic models exit the lab, our claim is not of immediate deployability but of testable prioritization: the bottleneck may not be more data but the right data: dense, long-horizon, ecologically valid trajectories. Robot Fostering delivers exactly those signals---under rigorous governance and with modest resources---providing a replicable, ethically grounded path toward reliable domestic robotics.
\section*{Appendix A: Power \& toy simulation code (minimal)}
\subsection*{A.1 Power analysis (reproducible)}
\begin{lstlisting}[language=Python]
import numpy as np
from statsmodels.stats.power import TTestIndPower, FTestAnovaPower, NormalIndPower
Two-arm, continuous (t-test)
tt = TTestIndPower()
n_two = np.ceil(tt.solve_power(effect_size=0.5, alpha=0.05, power=0.80, alternative='two-sided'))
print("Two-arm (d=0.5): n per arm =", int(n_two)) # 64
Three-arm, continuous (ANOVA)
ft = FTestAnovaPower()
n_three_total = np.ceil(ft.solve_power(effect_size=0.25, alpha=0.05, power=0.80, k_groups=3))
print("Three-arm (f=0.25): total n =", int(n_three_total)) # 158
Two proportions: p1=0.50 vs p2=0.65
z = NormalIndPower()
p1, p2 = 0.50, 0.65
pooled_var = (p1*(1-p1) + p2*(1-p2)) / 2
es = (p2 - p1) / np.sqrt(pooled_var)
n_prop = np.ceil(z.solve_power(effect_size=es, alpha=0.05, power=0.80, alternative='two-sided'))
print("Two-proportion (0.50->0.65): n per arm =", int(n_prop)) # 167
\end{lstlisting}
\subsection*{A.2 Toy Monte Carlo (uncertainty and sensitivity)}
\begin{lstlisting}[language=Python]
import numpy as np
rng = np.random.default_rng(42)
Priors for phase success (mean near 0.85/0.60/0.70 with moderate concentration)
alphas = np.array([85, 60, 70]) # pseudo successes
betas = np.array([15, 40, 30]) # pseudo failures
N = 5000
p = rng.beta(alphas, betas, size=(N,3))
p_full = (p[:,0] * p[:,1] * p[:,2])
mean = p_full.mean()
lo, hi = np.quantile(p_full, [0.025, 0.975])
print(f"Full-chain success mean={mean:.3f}, 95% CI=({lo:.3f}, {hi:.3f})")
Sensitivity: +10% relative improvement in the Family phase
p2 = p.copy()
p2[:,1] = np.clip(p2[:,1]*1.10, 0, 1)
p_full2 = (p2[:,0]*p2[:,1]*p2[:,2])
print(f"Delta mean (Family +10% rel.) = {p_full2.mean() - mean:.3f}")
\end{lstlisting}


\begin{thebibliography}{99}
\bibitem{Xiao2025LIMI}
Y.~Xiao, et al., LIMI: Less Is More for Agency,'' \emph{arXiv preprint arXiv:2509.17567} (2025). \bibitem{IEEE2010PR2} IEEE Spectrum, Willow Garage Giving Away 11 PR2 Robots to Outside Research Labs,'' (2010).
\bibitem{ROS2010PR2}
ROS.org News, Software highlights from the first PR2 Beta conference call,'' (2010). \bibitem{UniBremenPR2} University of Bremen, Institute for Artificial Intelligence, PR2 Beta Program,'' project page (2010).
\bibitem{EC_AIAct}
European Commission, Artificial Intelligence Act: EU regulatory framework for artificial intelligence,'' official portal (accessed 2025). \bibitem{ISO42001} International Organization for Standardization, ISO/IEC 42001: Artificial intelligence management system,'' (2023).
\bibitem{ISO23894}
International Organization for Standardization, ISO/IEC 23894: Artificial intelligence --- Guidance on risk management,'' (2023). \bibitem{IEEE7001} IEEE Standards Association, IEEE 7001-2021: Transparency of Autonomous Systems,'' (2021).
\bibitem{BI_RoombaLeak}
Business Insider, Leaked test images from Roomba devices spark privacy concerns,'' (2022). \bibitem{Coulson2021} S.~Coulson, Citizen Sensing: An Action-Orientated Framework for Citizen Science,'' \emph{Frontiers in Communication} 6 (2021).
\bibitem{MS_Toolkit}
Making Sense Consortium, Citizen Sensing Toolkit,'' (2018). \bibitem{Bhadauria2011} D.~Bhadauria and V.~Isler, Robotic Data Mules for Collecting Data over Sparse Sensor Fields,'' \emph{Journal of Field Robotics} 28(3), 201--223 (2011).
\bibitem{Tekdas2008}
O.~Tekdas, V.~Isler, J.~Sweeney, and A.~Kumar, Using Mobile Robots to Harvest Data from Sensor Fields,'' \emph{Wireless Communications and Mobile Computing} 8(6), 711--722 (2008). \bibitem{BhadauriaWiley} D.~Bhadauria and V.~Isler, Robotic Data Mules...,'' Wiley Online Library entry (2011).
\bibitem{AIActTimeline}
ArtificialIntelligenceAct.eu, Implementation timeline,'' expert tracker (accessed 2025). \bibitem{ReutersAIAct} Reuters, EU sticks with timeline for AI rules; obligations for GPAI models from Aug 2025,'' (2025).
\bibitem{ISO13482}
International Organization for Standardization, ISO 13482: Robots and robotic devices --- Safety requirements for personal care robots,'' (2014). \bibitem{Alverhed2024} E.~Alverhed, et al., Autonomous last-mile delivery robots: a literature review,'' \emph{European Transport Research Review} 16, 9 (2024).
\bibitem{Pani2020}
A.~Pani, et al., Public acceptance of autonomous delivery robots in urban areas,'' \emph{Transportation Research Part C} 120, 102774 (2020). \bibitem{Starship2025} Starship Technologies, Public attitude data from campus deployments,'' press release (2025).
\bibitem{NIST_RMF}
NIST, Artificial Intelligence Risk Management Framework (AI RMF 1.0),'' (2023). \bibitem{GDPR35} European Data Protection Board, Guidelines on Data Protection Impact Assessment (DPIA) and determining whether processing is likely to result in a high risk'' for the purposes of Regulation 2016/679,'' WP248 rev.01 (2017). \bibitem{Puterman1994} M.~Puterman, \emph{Markov Decision Processes: Discrete Stochastic Dynamic Programming}. Wiley (1994). \bibitem{SuttonBarto2018} R.~S.~Sutton and A.~G.~Barto, \emph{Reinforcement Learning: An Introduction}. 2nd ed., MIT Press (2018). \bibitem{W3C_OWL2} W3C, OWL 2 Web Ontology Language Document Overview (Second Edition),'' W3C Recommendation (2012).
\bibitem{GebruDatasheets}
T.~Gebru et al., Datasheets for Datasets,'' \emph{Communications of the ACM} 64(12), 86--92 (2021). \bibitem{BenderDataStatements} E.~M.~Bender and B.~Friedman, Data Statements for NLP: Toward Mitigating System Bias and Enabling Better Science,'' \emph{TACL} 6, 587--604 (2018).
\bibitem{BenderParrots}
E.~M.~Bender, T.~Gebru, A.~McMillan-Major, and S.~Shmitchell, On the Dangers of Stochastic Parrots,'' \emph{FAccT} (2021). \bibitem{Behavior1K} Y.~Li et al., BEHAVIOR-1K: A Benchmark for Embodied AI with 1,000 Household Activities,'' \emph{CoRL} (2022).
\end{thebibliography}
\end{document}